\documentclass[11pt]{article}
\usepackage{coling2020}
\usepackage{times}
\usepackage{url}
\usepackage{latexsym}
\usepackage{hyperref}
\usepackage{graphicx}
\usepackage{multirow}
\usepackage{amsmath}
\usepackage{booktabs}
\usepackage{wrapfig}
\usepackage{lipsum}

\colingfinalcopy 

\title{Harnessing Cross-lingual Features to Improve Cognate Detection for Low-resource Languages}

\author{Diptesh Kanojia\textsuperscript{$\dagger$,$\clubsuit$,$\star$}, Raj Dabre\textsuperscript{$\diamond$}, Shubham Dewangan\textsuperscript{$\dagger$},\\ \textbf{Pushpak Bhattacharyya}\textsuperscript{$\dagger$},
\textbf{Gholamreza Haffari}\textsuperscript{$\star$}, and \textbf{Malhar Kulkarni}\textsuperscript{$\dagger$}\\
  \textsuperscript{$\dagger$}IIT Bombay, India,  \textsuperscript{$\diamond$}NICT, Japan \\
  \textsuperscript{$\clubsuit$}IITB-Monash Research Academy, India \\
  \textsuperscript{$\star$}Monash University, Australia \\
  {\tt \textsuperscript{$\dagger$}\{diptesh,pb,malhar\}@iitb.ac.in, \textsuperscript{$\diamond$}raj.dabre@nict.go.jp}\\
  {\tt \textsuperscript{$\dagger$}sdofficial1996@gmail.com, \textsuperscript{$\star$}gholamreza.haffari@monash.edu} \\}

\date{26-06-2020}

\begin{document}
\maketitle
\begin{abstract}

Cognates are variants of the same lexical form across different languages; for example ``fonema'' in Spanish and ``phoneme'' in English are cognates, both of which mean ``a unit of sound''. The task of automatic detection of cognates among any two languages can help downstream NLP tasks such as Cross-lingual Information Retrieval, Computational Phylogenetics, and Machine Translation. In this paper, we demonstrate the use of cross-lingual word embeddings for detecting cognates among fourteen Indian Languages. Our approach introduces the use of context from a knowledge graph to generate improved feature representations for cognate detection. We then evaluate the impact of our cognate detection mechanism on neural machine translation (NMT), as a downstream task. We evaluate our methods to detect cognates on a challenging dataset of twelve Indian languages, namely, Sanskrit, Hindi, Assamese, Oriya, Kannada, Gujarati, Tamil, Telugu, Punjabi, Bengali, Marathi, and Malayalam. Additionally, we create evaluation datasets for two more Indian languages, Konkani and Nepali\footnote{It is primarily spoken in Nepal, but is also adopted in the list of scheduled languages of the Republic of India.}. We observe an improvement of up to 18\% points, in terms of F-score, for cognate detection. Furthermore, we observe that cognates extracted using our method help improve NMT quality by up to 2.76 BLEU. We also release\footnote{\href{http://www.cfilt.iitb.ac.in/coling2020diptesh}{Link: Data, code and models}} our code, newly constructed datasets and cross-lingual models publicly.
\end{abstract}

\section{Introduction}
\label{sec:intro}

India is a multilingual, multi-script country with 22 scheduled languages and 12 written script forms primarily belonging to 6 different language families. More than a billion people use these languages as their first language. A significant amount of news and information is found on the web in these languages, which is inaccessible to people of other regions within the country. Most of the Indian language texts found online have several words that have originated from Sanskrit, Persian, and English. While, in many cases, one might argue that such occurrences do not belong to an Indian language, the frequency of such usage indicates a wide acceptance of these foreign language words as Indian language words. In numerous cases, these words also are morphologically altered as per the Indian language morphological rules to generate new variants of existing words. Detection of such variants or `Cognates' across languages helps Cross-lingual Information Retrieval (CLIR)~\cite{makin2008experiments,meng2001generating}, Machine Translation (MT)~\cite{kondrak2005cognates,kondrak2003cognates,al1999statistical}, and Computational Phylogenetics~\cite{rama2018automatic}. Cognates are etymologically related words across two languages~\cite{crystal2011dictionary}. However, NLP applications are concerned with the set of cognate words which have similarities in their spelling and their meaning. For example, the French and English word pair, \textit{Libert{\'e} - Liberty}, reveals itself to be a true cognate through orthographic similarity. In some cases, similar words have a common meaning only in some contexts; such words are called partial cognates. For example, the word \textit{``police''} in French can translate to \textit{``police''}, \textit{``policy''} or \textit{``font''}, depending on the context\footnote{Cognates can also exist in the same language. Such word pairs/sets are commonly referred to as \textit{doublets}.}.\blfootnote{
    %
    %
    %
    %
    \hspace{-0.65cm}  
    This work is licensed under a Creative Commons Attribution 4.0 International Licence. Licence details:
    \url{http://creativecommons.org/licenses/by/4.0/}.
    %
    %
} Manual detection of such cognate sets requires a human expert with a good linguistic background in multiple languages. Moreover, manual annotation of cognate sets is a costly task in terms of time and human effort. 

The task of cognate detection across languages requires one to detect word pairs which are etymologically related, and carry the same meaning. Previous approaches to the task use orhtographic~\cite{ciobanu2014automatic}, phonetic~\cite{rama2016siamese} and semantic~\cite{kondrak2001identifying} features. However, these methods have a limitation since they do not take into consideration the notion of semantic similarity across languages.
A key question that we try to answer in this paper is,
\begin{quote}
\textit{``Can semantic information be leveraged from Cross-lingual models to improve cognate detection amongst low-resource languages?''}
\end{quote}
We hypothesize that utilizing cross-lingual features by employing existing resources such as wordnets and cross-lingual embeddings should help improve cognate detection. In this paper, we utilize the semantic information from cross-lingual word embeddings. Cross-lingual word embeddings can be obtained by training monolingual embeddings for individual languages and then projecting them in a shared space using a bilingual dictionary. In the absence of such a bilingual dictionary for low-resource languages, adversarial training can be used over identical words to generate the projections. We build cross-lingual models for thirteen language pairs with Hindi as the source (L1) and thirteen target Indian languages (L2). We use the context information from a knowledge graph to build the context dictionaries for each pair. The cross-lingual models help us obtain embeddings for the word-pair and the respective context dictionaries, from a shared space. We hypothesize that using this approach should provide a more accurate semantic measure for the detection of cognate pairs. The use of orthographic and phonetic similarity-based methods to perform the same task provides us with baselines for a comparative evaluation.

A motivation to investigate this task for low-resource Indian languages stems from the fact that most of the Indian languages borrow cognates or ``loan words'' from the Sanskrit language. It is, for the most part, considered a historical antecedent of almost all the Indian languages. Indo-Aryan languages like Hindi, Bengali, Gujarati, Punjabi borrow from Sanskrit. They borrow many lexical forms and language properties from Sanskrit. Dravidian languages are highly agglutinative and morphologically rich like Sanskrit, which makes them tough to parse computationally. Marathi and Hindi suffer from the same ailment even though Hindi is not considered as agglutinative as Marathi, but it does exhibit compounding\footnote{Compounding means when two or more words or signs are joined to make a longer word or sign.} which makes it, yet again, difficult to parse for CLIR and MT systems, and to detect cognates based solely on orthographic similarity. Given that CLIR and MT are usually based on a full-form lexicon, one of the possible issues in the generation of cognates concerns the similarity of words in their root form vs the similarity in their lexical form. For example, the Sanskrit word ``\textit{matra}'' and the English word ``\textit{Mother}'' are known cognates from the Proto-Indo-European language family where the root and the meaning are identical, but the lexical form is considerably different. Our approach handles such cases by inculcating the sub-word information while building the embeddings and helps reduce the out-of-vocabulary (OOV) words, which have proven to be a challenge for well-established CLIR systems \cite{udupa2009they}.

This paper is organized as follows. Section \ref{sec:rw} briefly describes the previous work done in the area of automatic cognate detection. Section \ref{sec:des} describes the dataset source, our additions to it, and the experimental setup. Section \ref{sec:approaches} presents the approaches used in terms of feature sets and classification methodologies. The results obtained are described in Section \ref{sec:results} along with a discussion on the qualitative analysis of our output. Section \ref{sec:conc} concludes this article with possible future work in the area.

\section{Related Work}
\label{sec:rw}

The two main existing approaches for the detection of cognates belong to the \textit{generative} and \textit{discriminative} paradigms. The first set of approaches is based on the computation of a similarity score between potential candidate pairs. This score can be based on orthographic similarity \cite{jager2017using,melamed1999bitext,mulloni2006automatic}, phonetic similarity \cite{rama2016siamese,list2012lexstat,kondrak2000new}, or a distance measure with the scores learned from an existing parallel set \cite{mann2001multipath,tiedemann1999automatic}. The discriminative paradigm uses standard approaches to machine learning, which are based on (1) extracting features, \textit{e.g.,} character n-grams, and (2) learning to predict the transformations of the source word needed to \cite{jiampojamarn2010integrating,frunza2009identification}. 

Cognate Detection has been explored vastly in terms of classification methodologies. Previously, \newcite{rama2016siamese} employ a Siamese convolutional neural network to learn the phonetic features jointly with language relatedness for cognate identification, which was achieved through phoneme encodings. \newcite{jager2017using} use SVM for phonetic alignment and perform cognate detection for various language families. Various works on orthographic cognate detection usually take alignment of substrings within classifiers like SVM \cite{ciobanu2014automatic,ciobanu2015automatic} or HMM \cite{bhargava2009multiple}. \newcite{ciobanu2014automatic} employ dynamic programming based methods for sequence alignment. \newcite{kanojia2019cognate} perform cognate detection for some Indian languages, but a prominent part of their work includes \textit{manual verification and segratation} of their output into cognates and non-cognates. \newcite{kanojia2019utilizing} utilize recurrent neural networks to harness the character sequence among cognates and non-cognates for Indian languages, but employ monolingual embeddings for the task. \newcite{dijkstra2010cross} show how cross-linguistic similarity of translation equivalents affects bilingual word recognition, even in tasks manually performed by humans. They discuss how the need for recognizing semantic similarity arises for non-identical cognates, based on the reaction time from human annotators. Similarly, \newcite{merlo-andueza-rodriguez-2019-cross} show that cross-lingual models exhibit the semantic properties of for bilingual lexicons despite their structural simplicities, which leads us to perform our investigation for low-resource Indian languages. \newcite{uban-etal-2019-studying} discuss the semantic change in languages by studying the change in cognate words across Romance languages using cross-lingual similarity. All of the previous approaches discussed above, lack the use of an appropriate cross-lingual similarity-based measure and do not work well for Indian languages as shown in this work. This paper discusses the quantitative and qualitative results using our approach and then, applies our output to different neural machine translation architectures.

\renewcommand{\tabcolsep}{3pt}

\begin{table*}[ht]
\centering
\resizebox{0.95\textwidth}{!}{%
\begin{tabular}{|c|c|c|c|c|c|c|c|c|c|c|c|c|c|}
\hline
\textbf{Language Pair}     & \textbf{Hi-Bn} & \textbf{Hi-Gu} & \textbf{Hi-Mr} & \textbf{Hi-Pa} & \textbf{Hi-Sa} & \textbf{Hi-Ml} & \textbf{Hi-Ta} & \textbf{Hi-Te} & \textbf{Hi-As} & \textbf{Hi-Kn} & \textbf{Hi-Or} & \textbf{Hi-Ne*}  & \textbf{Hi-Ko*} \\ \hline

\textbf{Cognates}          & $15312$          & $17021$          & $15726$          & $14097$          & $21710$          & $9235$           & $3363$           & $936$            & $3478$           & $4103$          & $11894$          & $2560$          & $11295$          \\ \hline
\textbf{Non-Cognates}             & $16119$        & $15057$        & $15983$        & $15166$        & $23029$        & $8976$        & $4005$         & $1084$        & $4101$        & $3810$        & $13027$          & $1918$          & $9826$        \\ \hline

\end{tabular}
}
\caption{Number of cognates and non-cognates for each language pair in the dataset. Hi-Ne* and Hi-Ko* were generated via replicating their approach (Kanojia et. al., 2020).}
\label{tab:d2stats}
\end{table*}
\vspace{-0.5cm}
\begin{table*}[h]
\centering
\resizebox{0.95\textwidth}{!}{%
\begin{tabular}{|c|c|c|c|c|c|c|c|c|c|c|c|c|c|c|}
\hline
\textbf{Language}      & \textbf{Hi} & \textbf{Bn} & \textbf{Gu} & \textbf{Mr} & \textbf{Pa} & \textbf{Sa} & \textbf{Ml} & \textbf{Ta} & \textbf{Te} & \textbf{Ne} & \textbf{As} & \textbf{Kn} & \textbf{Ko} & \textbf{Or} \\ \hline
\textbf{Corpus Size}   & 48142K      & 1564K       & 439K        & 520K        & 505K        & 553K        & 495K        & 909K        & 1023K       & 706K        & 504K        & 159K        & 214K        & 744K        \\ \hline
\textbf{STTR (n=1000)} & 0.5821      & 0.5437      & 0.4587      & 0.6108      & 0.4314      & 0.5350      & 0.7339      & 0.6411      & 0.4950      & 0.4883      & 0.5968      & 0.5338      & 0.5614      & 0.4160      \\ \hline
\end{tabular}
}
\caption{Corpus Statistics where corpus size is the approximate number of lines, and STTR is the moving average type-token ratio on a windows of 1000 sentences.}
\label{tab:embedstats}
\end{table*}
\renewcommand{\tabcolsep}{6pt}
\vspace{-0.5cm}

\section{Dataset and Experimental Setup}
\label{sec:des}

In this section, we describe our primary dataset for the cognate detection task. We also describe the datasets used for building cross-lingual word embedding models, and the parallel corpora used for the Neural Machine Translation (NMT).
\begin{wrapfigure}{r}{0.50\columnwidth}
  \begin{center}
    \vspace{-14pt}
    \includegraphics[width=0.49\textwidth]{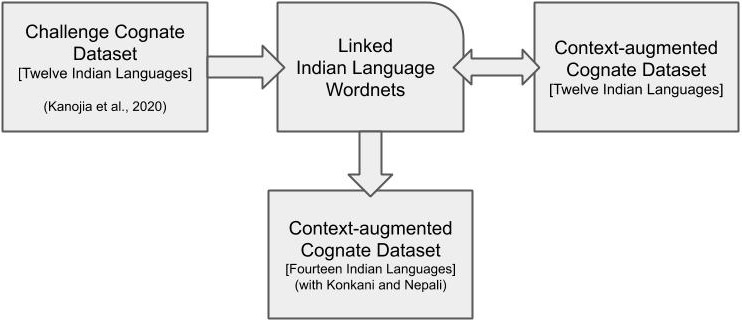}
    \vspace{-20pt}
  \end{center}
  \caption{Dataset Augmentation with Context and Two Language Pairs using IndoWordnet.}
  \label{fig:dataset}
\end{wrapfigure}
For our experiments, we use the publicly released challenge dataset~\cite{kanojia2020challenge} of cognates. This dataset provides labelled cognate and non-cognate pairs for twelve Indian languages namely, Sanskrit (Sa), Hindi (Hi), Assamese (As), Oriya (Or), Kannada (Kn), Gujarati (Gu), Tamil (Ta), Telugu (Te), Punjabi (Pa), Bengali (Bn), Marathi (Mr), and Malayalam (Ml). We reproduce their approach to add two more languages, Konkani (Ko) and Nepali (Ne), to this dataset. For building context dictionaries, we use linked Indian language wordnets~\cite{bhattacharyya2017indowordnet} and concatenate the concept definition and example sentences. We remove stop words from the context dictionaries and append them with their respective word pairs. The lexical overlap between the language pairs ranges from 13\% (for Hi-Te) to only 23\% (Hi-Mr). Figure \ref{fig:dataset} shows an accurate description of the dataset creation process. The cognate dataset statistics are described in Table \ref{tab:d2stats}.

\subsection*{Monolingual Corpora for Word Embeddings}

The dataset for training cross-lingual models is obtained from various sources. Word embeddings require a large quantity of monolingual corpora for efficient training of a usable model with high accuracy. We extract corpora for these fourteen Indian Languages from various sources and collect them in a single repository. We extract Wikimedia dumps\footnote{\href{https://dumps.wikimedia.org/}{Link: Wikimedia Dumps; as on April 22, 2020}} for all languages and add Indian Language Corpora Initiative (ILCI) corpora~\cite{jha-2010-tdil} for these languages to each of them. For Hindi, Marathi, Nepali, Bengali, Tamil, and Gujarati we add crawled corpus of film reviews and news websites\footnote{\href{https://github.com/goru001?tab=repositories}{Link: Additional Monolingual Corpus}} to their corpus. For Hindi, we also add HinMonoCorp 0.5~\cite{bojar2014hindencorp} to our corpus adding approximately 44 million sentences. For Sanskrit, we download a raw corpus of proses\footnote{\href{http://sanskrit.jnu.ac.in/currentSanskritProse/}{Link: JNU Sanskrit Proses Corpus}} and add it to our corpus. Training corpus statistics (approximate number of total lines) are shown in Table \ref{tab:embedstats}.

\subsection*{Parallel Corpora for NMT}

To validate the application of cognates for the Machine Translation task, we choose the Neural Machine Translation setting and use the Indian Languages Corpora Initiative (ILCI) Phase 1 corpus. This corpus contains approximately 50K parallel sentences across 11 languages (English and 10 Indian Languages), from health and tourism domains. For every language pair, the parallel corpus was split up into a training set of 46,277 sentences, a test set of 2000 sentences and development set of 500 sentences. The train, test and development splits were ensured to be parallel across all language pairs involved. The language pair intersection for our cognate detection work and this parallel corpus limited our MT experimentation to the following languages namely, Hindi (Hi), Punjabi (Pa), Bengali (Bn), Gujarati (Gu), Marathi (Mr), Tamil (Ta), Telugu (Te) and Malayalam (Ml). We keep Hi as the source and remaining languages as the target languages for our experiments. We describe the experimental setup for our task below.

\subsection{Unicode Offset based Transliteration}

Indian languages use different scripts, and lexical similarity-based metrics cannot be directly used on any text for character matching. For standardization, we choose to convert any other script to the Devanagari script. We perform Unicode transliteration using Indic NLP Library\footnote{\href{https://anoopkunchukuttan.github.io/indic\_nlp\_library/}{Link: Indic NLP Library}} to convert scripts for Bn, As, Or, Gu, Pa, Ml, Ta, Kn and Te to Devanagari for standardization. Hi, Mr, Ko, Ne, and Sa are already based on the Devanagari script. We perform this for script transliteration for both the cognate dataset (Table \ref{tab:d2stats}) and the corpus (Table \ref{tab:embedstats}). We describe the creation of cross-lingual word embeddings below.

\subsection{Cross-lingual Word Embedding Methodologies}

Using the monolingual corpora described above, we build monolingual word embeddings using the FastText library\footnote{\href{https://github.com/facebookresearch/fastText}{Link: FastText - GitHub}}~\cite{bojanowski2017enriching} since it takes sub-word information into account, which is beneficial for a task such as ours where sub-words play an important role, and spelling variations can lead to different meanings. We do not use BERT~\cite{devlin2018bert}, ELMo~\cite{peters2018deep}, or M-BERT~\cite{pires-etal-2019-multilingual} for word embeddings as their pre-trained models are not trained on transliterated corpora. We choose FastText to train Skipgram word embedding models (100 dimensions) for each language using the following hyperparameters - 15 epochs with 0.1 as the learning rate. We use two characters (bi-gram) as the size of each sub-word for capturing the maximum number of sub-words. 

We use three different methodologies for training the cross-lingual word embedding models on all the language pairs with Hindi as a pivot language (Hi-Mr, Hi-Bn and so on). The \textbf{first methodology} uses the supervised method named MUSE~\cite{conneau2017word}\footnote{\href{https://github.com/facebookresearch/MUSE}{Link: MUSE - GitHub}} which utilizes a manually curated bilingual lexicon\footnote{\href{http://www.cfilt.iitb.ac.in/Downloads.html}{Link: Bilingual Lexicon}} for alignments. We use Hindi as a pivot language due to the ease of computation and availability of resources (Corpora and WordNet size). We use the monolingual models described above and train 13 cross-lingual word embedding models (thirteen language pairs over 100 dimensions) using this approach. 

The \textbf{second cross-lingual methodology} uses VecMap~\cite{artetxe2018acl}, which utilizes the monolingual models created above. VecMap uses an optional normalization feature while it builds the mappings between any two monolingual models. It performs orthogonal transformation and maps semantically related words, similar to MUSE, which was used in our first approach for building cross-lingual models. Additionally, it also reduces the dimensions of the embeddings models, which, is optional. We train it using the same hyperparameters as described above, for consistency while evaluating. We used the supervised approach for training these models as well, and the training dictionary was similar to the one provided to the MUSE method. We obtain thirteen models, one for each language pair, using VecMap. The \textbf{third methodology} utilizes contextual embeddings which have shown to outperform the conventional word embeddings based models for many tasks~\cite{devlin2018bert}. We choose the most recent methodology for building a single cross-lingual model for all the languages. XLM-R~\cite{conneau2019unsupervised} uses previously proposed approaches of XLM~\cite{lample2019cross} and RoBERTa~\cite{liu2019roberta} to attain a very high performing cross-lingual model, especially for low-resource languages. We use our transliterated corpora described above and concatenate it into a single large corpus required for training the model. We then use the unsupervised training method of XLM-R and train a model over six days and a couple more hours with a reduced batch size.



\textit{To put it more concisely, we trained cross-lingual models using three different methodologies (MUSE, VecMap and XLM-R)} where the cross-lingual mapping obtained for MUSE and VecMap were generated via the monolingual embeddings, as described above. We obtained thirteen models using each of these two methods. A single cross-lingual model was, however, trained using XLM-R and used for the third cross-lingual approach whose training methodology has been described above. We utilize the last layer from the XLM-R model to generate representations for each token.

\section{Approaches}
\label{sec:approaches}

We use various approaches to perform the cognate detection task \textit{viz.} baseline cognate detection approaches like orthographic similarity-based, phonetic similarity-based, phonetic vectors with Siamese-CNN based proposed by \newcite{rama2016siamese}, and Recurrent neural network-based approach proposed by \newcite{kanojia2019utilizing}. We use the same hyperparameters and architectures, as discussed in these papers. We describe each of these feature sets in this section.

\subsection{Weighted Lexical Similarity (WLS)}

The Normalized Edit Distance (NED) approach computes the edit distance~\cite{nerbonne1997measuring} for all word pairs in our dataset. Each of the operations has unit cost (except that substitution of a character by itself has zero cost), so NED is equal to the minimum number of operations to transform `word a' to `word b'. We use a similarity score provided by NED, which is calculated as (1 - NED Score). We combine NED with q-gram distance~\cite{shannon1948mathematical} for a better similarity score. The q-grams (`n-grams') are simply substrings of length q. This distance measure has been applied previously for various spelling correction approaches~\cite{owolabi1988fast,kohonen1978very}. \newcite{kanojia2019utilizing} propose this metric and we replicate it to generate features for their baseline approach. For any word pair with words $p$ and $q$, it is as follows:
\begin{equation}
\begin{aligned}
WLS_{pq} = (NED_{pq} * 0.75) + (QD_{pq} * 0.25)
\end{aligned}
\end{equation}
Now that this approach can be used to compute a score between each word pair, we use it to find two scores, which are used as features - `word-pair similarity' and `contextual similarity'. Each candidate word-pair generates a score \textit{i.e.}, score1, and the average of scores among all words in the context dictionary generates another score \textit{i.e.}, score2, which are normalized as follows:
\begin{equation}
\begin{aligned}
S_{1} = score1 / \left(score1 + score2\right)\\
S_{2} = score2 / \left(score1 + score2\right)
\end{aligned}
\end{equation}

We use $S_{1}$ and $S_{2}$ as features for this orthographic similarity-based baseline approach. 

\subsection{Phonetic Vectors and Similarity (PVS)}

The IndicNLP Library provides phonetic features based vector for each character in various Indian language scripts. We utilize this library to compute a feature vector for each word by computing an average over character vectors. We compute vectors for both words in the candidate cognate pairs ($PV_S$ and $PV_T$) and also compute contextual vectors ($PCV_S$ and $PCV_T$) by averaging the vectors for all the context dictionaries on each side (source and target), generating a total of four vectors. We also calculate the cosine similarity among $PV_S$ and $PV_T$, and among $PCV_S$ and $PCV_T$ to generate two similarity scores ($P_{S1}$, and $P_{S2}$) which are normalized using (2) and, additionally, used as features during classification. It should be noted that using phonetic vectors and their similarity scores has already been proposed in the previous literature~\cite{rama2016siamese} for a cognate detection task, and we do not claim this approach to be our novel contribution. 

\subsection{Cross-lingual Vectors \& Similarity}

As described above, we train cross-lingual embedding models by aligning two disjoint monolingual vector spaces through linear transformations, using a small bilingual dictionary for supervision~\cite{doval2018improving,artetxe-etal-2017-learning}. The first two approaches for training cross-lingual methods use this dictionary for supervision. In our novel approach, we propose the use of vectors from the cross-lingual embedding models trained on Indian language pairs. We obtain vectors for word-pairs ($WV_S$ and $WV_T$) and averaged context vectors ($CV_S$ and $CV_T$) for the context dictionary, to create feature sets. We obtain vectors for each candidate pair and their context using all the three cross-lingual methodologies.

Additionally, we use angular cosine similarity~\cite{cer2018universal} scores for word pairs and their contexts. Angular similarity distinguishes nearly parallel vectors much better as small changes in vector values yield considerable distances. For each word pair vector and its context vectors, we compute the `word-pair similarity' and `contextual similarity'. We use \textit{arccos} to obtain angular cosine similarity (asim) among vectors `u' and `v', as shown below:\\
\begin{equation}
\begin{aligned}
asim(u,v) = \left( 1 - arccos \left(\frac{u.v}{\|u\| \|v\|} \right)/ \pi \right)
\end{aligned}
\end{equation}
Each candidate word-pair generates a score \textit{i.e.}, score1, and the average of scores among all words in the context dictionary generates another score \textit{i.e.}, score2, which are also normalized using (2).

\subsection{Classification Methodology}

\begin{figure*}[ht!]
    \centering
    \includegraphics[width=0.80\textwidth]{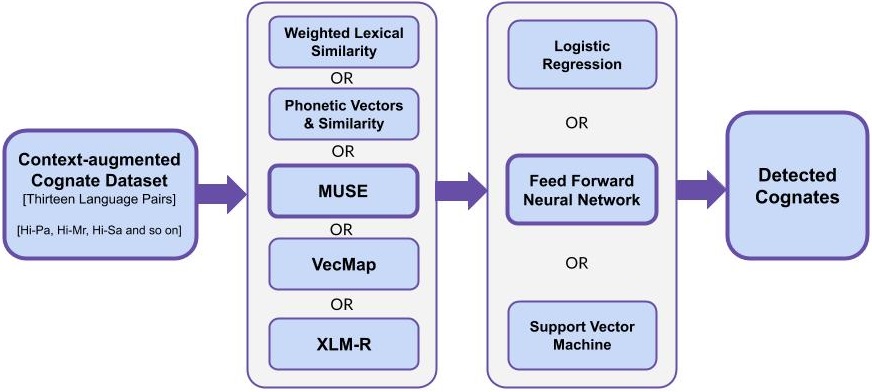}
    \caption{Cognate Detection task with different feature sets and classification approaches.}
    \label{fig:arch}
\end{figure*}

We pose the task of detecting cognates as a binary classification problem. We employ both classical machine learning-based models and a simple feed-forward neural network. To compare our work with the previously proposed approaches, we replicate the best-reported systems from \newcite{rama2016siamese} \textit{i.e.,} Siamese Convolutional Neural Network with phonetic vectors as features and also replicate \newcite{kanojia2019utilizing}'s approach which uses a Recurrent Neural Network architecture with a weighted lexical similarity (WLS) as a feature set. The input to our classifiers is the feature sets described above for each candidate pair. The candidates are the complete data described in Table \ref{tab:d2stats}. Cognates from Table \ref{tab:d2stats} are labelled positive, and non-cognates are labelled negative. \textit{We perform 5-fold stratified cross-validation, which divides the data into train and test folds, randomly.} An architecture diagram for our classification approach is shown in Figure \ref{fig:arch}.

Among the classical machine learning models, we use Support Vector Machines (SVM) and Logistic Regression (LR). We experiment with the use of both linear SVMs and kernel SVMs (Gaussian and Polynomial). We perform a grid-search to find the best hyper-parameter value for $C$ over the range of 0.01 to 1000. We deploy the Feed Forward Neural Network (FFNN) with one hidden layer. We perform cross-validation with different settings for activation function (tanh, hardtanh, sigmoid and relu) and the hidden layer dimension in the network (30, 50, 100, and 150). We use binary cross-entropy as the optimization algorithm. Finally, we choose the hyper-parameter configuration with the best validation accuracy. We train the model with the selected configuration with an initial learning rate of 0.4, and we halve the learning rate when the error on the validation split increases. We stop the training once the learning rate falls below $0.001$. We perform our experiments with the feature sets (Orthographic (WLS), Phonetic (PVS), and three different cross-lingual embeddings based feature sets) described above for all the thirteen language pairs. We also perform an ablation test with various feature sets and report the results for the best feature combination in the next section. The results of our classification task can be seen in Table \ref{tab:results} and are discussed in the next section, in detail.

\subsection{Cognate-aware Neural Machine Translation (NMT) Task}

For the NMT task, we use the OpenNMT-Py toolkit~\cite{klein2017opennmt} to perform our experiments. We use a Bidirectional RNN Encoder-Decoder architecture with attention \cite{bahdanau2014neural}. We choose three stacked LSTM \cite{lstm} layers in the encoder and decoder. The hidden-size of the model was 500 units. We optimize using stochastic gradient descent at an initial learning rate of 1, and a batch-size of 1024 units. Training is done for 150,000 steps of which the initial 8,000 steps are for learning rate warm-up. We use Byte-pair encoding (BPE)~\cite{sennrich2015neural} merge operations, initially, in an endeavour to find the best baseline model with an optimal number of merge operations. We observe that performing 2500 merge operations provided us with best BLEU~\cite{papineni2002bleu} scores, for most of the language pairs. We report the best results here, and a complete set of merge operation results in the supplementary material. We call this the NMT-BPE Baseline. 

To validate our hypothesis that our approach can help the NMT task, we \textit{inject the cognates detected using our approach} to the parallel corpus for their respective language pairs, as single word sentences. Lexical Dictionaries have previously been used to improve the MT task~\cite{arthur2016incorporating,han2019explicitly}. However, a decent improvement in their BLEU scores is observed when their lexicon sizes are approximately around 1M tokens~\cite{arthur2016incorporating}. Our detected cognate list size varies from 930 cognates (Hi-Te) to 15834 (Hi-Mr). Due to the addition of more parallel instances to the corpus, the vocabulary size for NMT increases. Hence, we experiment further by varying the BPE merges, in a close range, to the optimal merge point obtained earlier. We report the results of the best optimal merge setting, for both NMT-BPE Baseline model and the cognate injected NMT-BPE model, in the section below. A more detailed set of results for all the merge operations is available in the supplementary material.

\section{Results and Discussion}
\label{sec:results}

\renewcommand{\tabcolsep}{3pt}
\begin{table}[]
\centering
\resizebox{\textwidth}{!}{%
\begin{tabular}{c|ccc|ccc|ccc|ccc|ccc|ccc|ccc}
\toprule
 & \multicolumn{9}{c|}{Baseline Approaches} & \multicolumn{9}{c|}{Cross-lingual Embeddings based Approaches} & \multicolumn{3}{c}{Best Combination} \\ \midrule
\multirow{2}{*}{LP} & \multicolumn{3}{c|}{WLS w/ FFNN} & \multicolumn{3}{c|}{\begin{tabular}[c]{@{}c@{}}PVS \\ w/\\ Siamese CNN\\ (Rama, 2016)\end{tabular}} & \multicolumn{3}{c|}{\begin{tabular}[c]{@{}c@{}}WLS w/ RNN\\ (Kanojia et al.,\\  2019)\end{tabular}} & \multicolumn{3}{c|}{\begin{tabular}[c]{@{}c@{}}XLM-R \\ w/ FFNN\end{tabular}} & \multicolumn{3}{c|}{\begin{tabular}[c]{@{}c@{}}MUSE \\ w/ FFNN\end{tabular}} & \multicolumn{3}{c|}{\begin{tabular}[c]{@{}c@{}}VecMap \\ w/ FFNN\end{tabular}} & \multicolumn{3}{c}{\begin{tabular}[c]{@{}c@{}}MUSE + WLS\\ w/\\ FFNN\end{tabular}} \\ \cline{2-22} 
 & P & R & F & P & R & F & P & R & F & P & R & F & P & R & F & P & R & F & P & R & \multicolumn{1}{c}{F} \\ \midrule
Hi-Bn & 0.51 & 0.28 & 0.36 & 0.68 & 0.62 & 0.65 & 0.67 & 0.69 & 0.68 & 0.81 & 0.76 & \textbf{0.78} & 0.77 & 0.75 & 0.76 & 0.72 & 0.74 & 0.73 & 0.80 & 0.75 & 0.77 \\
Hi-As & 0.48 & 0.26 & 0.34 & 0.72 & 0.71 & 0.71 & 0.72 & 0.70 & 0.71 & 0.70 & 0.72 & 0.71 & 0.80 & 0.75 & \textbf{0.77} & 0.74 & 0.73 & 0.73 & 0.84 & 0.75 & \textbf{0.79} \\
Hi-Or & 0.51 & 0.30 & 0.38 & 0.65 & 0.58 & 0.61 & 0.66 & 0.58 & 0.62 & 0.65 & 0.61 & 0.63 & 0.72 & 0.68 & \textbf{0.70} & 0.67 & 0.70 & 0.68 & 0.81 & 0.69 & \textbf{0.75} \\
Hi-Gu & 0.43 & 0.16 & 0.23 & 0.70 & 0.65 & 0.67 & 0.81 & 0.71 & 0.76 & 0.80 & 0.73 & 0.76 & 0.80 & 0.84 & \textbf{0.82} & 0.77 & 0.74 & 0.75 & 0.83 & 0.85 & \textbf{0.84} \\
Hi-Ne & 0.50 & 0.16 & 0.24 & 0.72 & 0.84 & 0.78 & 0.78 & 0.73 & 0.75 & 0.75 & 0.75 & 0.75 & 0.86 & 0.83 & \textbf{0.84} & 0.78 & 0.73 & 0.75 & 0.86 & 0.83 & \textbf{0.84} \\
Hi-Mr & 0.51 & 0.20 & 0.29 & 0.70 & 0.68 & 0.69 & 0.74 & 0.70 & 0.72 & 0.76 & 0.71 & \textbf{0.73} & 0.70 & 0.73 & 0.71 & 0.71 & 0.71 & 0.71 & 0.72 & 0.73 & 0.72 \\
Hi-Ko & 0.47 & 0.24 & 0.32 & 0.63 & 0.63 & 0.63 & 0.63 & 0.59 & 0.61 & 0.66 & 0.58 & 0.62 & 0.69 & 0.73 & \textbf{0.71} & 0.61 & 0.60 & 0.60 & 0.70 & 0.75 & \textbf{0.72} \\
Hi-Pa & 0.28 & 0.17 & 0.21 & 0.51 & 0.44 & 0.47 & 0.76 & 0.72 & 0.74 & 0.75 & 0.71 & 0.73 & 0.83 & 0.78 & \textbf{0.80} & 0.71 & 0.74 & 0.72 & 0.83 & 0.78 & \textbf{0.80} \\
Hi-Sa & 0.34 & 0.19 & 0.24 & 0.55 & 0.51 & 0.53 & 0.73 & 0.71 & 0.72 & 0.75 & 0.70 & 0.72 & 0.77 & 0.76 & \textbf{0.76} & 0.73 & 0.71 & 0.72 & 0.80 & 0.77 & \textbf{0.78} \\
Hi-Ml & 0.49 & 0.20 & 0.28 & 0.59 & 0.66 & 0.62 & 0.66 & 0.66 & 0.66 & 0.72 & 0.63 & 0.67 & 0.76 & 0.71 & \textbf{0.73} & 0.69 & 0.71 & 0.70 & 0.77 & 0.71 & \textbf{0.74} \\
Hi-Ta & 0.22 & 0.19 & 0.20 & 0.49 & 0.58 & 0.53 & 0.49 & 0.58 & 0.53 & 0.63 & 0.51 & 0.56 & 0.72 & 0.68 & \textbf{0.70} & 0.66 & 0.72 & 0.69 & 0.72 & 0.70 & \textbf{0.71} \\
Hi-Te & 0.18 & 0.15 & 0.16 & 0.60 & 0.71 & 0.65 & 0.62 & 0.71 & 0.66 & 0.65 & 0.70 & 0.67 & 0.70 & 0.72 & \textbf{0.71} & 0.67 & 0.67 & 0.67 & 0.73 & 0.72 & \textbf{0.72} \\
Hi-Kn & 0.19 & 0.18 & 0.18 & 0.54 & 0.60 & 0.57 & 0.58 & 0.60 & 0.59 & 0.60 & 0.58 & 0.59 & 0.69 & 0.73 & \textbf{0.71} & 0.65 & 0.64 & 0.64 & 0.70 & 0.73 & \textbf{0.71} \\ \bottomrule
\end{tabular}%
}
\vspace{-6pt}
\caption{Results of the cognate detection task, in terms of weighted F-scores (5-fold) with baseline features and previous approaches, and our approaches using Cross-lingual similarity based features, for all the language pairs (LP).}
\label{tab:results}
\vspace{-11pt}
\end{table}
\renewcommand{\tabcolsep}{6pt}

From Table \ref{tab:results}, among the baseline approaches, we observe high precision but very low recall scores when Weighted Lexical Similarity (WLS) based features are used. In fact, for language pairs which contain the Dravidian languages (Hi-Ml, Hi-Ta, Hi-Te, and Hi-Kn), even the precision scores are observed to be very low. The classifiers are not able to predict a significant amount of positively labelled cognate pairs, correctly. Even simple lexical variants such as ``\textit{Aag} (Fire)'' (Hindi) and ``\textit{Agni} (Fire)'' (Telugu) were classified incorrectly, as non-cognates. Phonetic vectors paired with a Siamese CNN~\cite{rama2016siamese}, however, mitigate such misclassifications and are shown to perform well with much higher recall, for all the language pairs. \newcite{kanojia2019utilizing}'s approach, however, outperforms the phonetic vectors based approach. We observe marginal improvements in F-scores for almost all the language pairs (except Hi-Ko and Hi-Ne) when their RNN based approach is used. As for our approaches, SVM and Logistic Regression based classification methodologies were consistently outperformed by the FFNN method. Hence, we report precision (P), recall (R), and F-scores (F) for only FFNN based approaches in Table \ref{tab:results}.

Our cross-lingual similarity-based approaches, however, significantly outperform all the baseline approaches. We observe a stark improvement in both precision and recall scores for all the language pairs. The cross-lingual approach, which uses the vectors from VecMap based models, fails to outperform both MUSE and XLM-R based models. XLM-R model exclusively achieves the best f-score for two language pairs (Hi-Bn and Hi-Mr). We believe its performance can be attributed to the closeness of the language pairs as they belong to the same language family (Indo-Aryan). Moreover, XLM-R is a transformer architecture-based model which requires relatively larger corpora sizes and a decent amount of corpus was available to build word embedding models for these target languages (Table \ref{tab:embedstats}). The cross-lingual models built above are used to provide vectors for calculating the similarity between words and contexts, bringing in the notion of semantic similarity for the task of cognate detection. Please note that by the definition of cognates, they are semantically similar despite the lexical variance. We observe that MUSE based feature representations paired with FFNN, obtain the best F-scores. This observation stands true even when the target language belongs to the Dravidian language family, where our baseline approaches lack severely in performance. For example, ``\textit{mkarand-maKarantam} (pollen)'' (Hi-Ta), a cognate pair was classified correctly only using the MUSE based approach.

Additionally, we perform an ablation test with our feature sets for further experimentation. We observe that the combination of WLS and vectors from the MUSE model performs even better. An improvement is observed for eight language pairs out of thirteen ranging from 1\% point (Hi-Ko, Hi-Ml, Hi-Ta, Hi-Te) to 5\% points (Hi-Or). It should be noted that this is the only combination where no degradation in performance was observed for any language pair and hence, is reported in Table \ref{tab:results}. Any other combination (MUSE + VecMap, MUSE + XLM-R, MUSE + PVS, and so on) degrades the performance of the cognate detection task, on at least one language pair. 

\renewcommand{\tabcolsep}{0.2cm}
\begin{wraptable}{r}{10.6cm}
\centering
\resizebox{0.65\textwidth}{!}{%
\begin{tabular}{@{}cccccccc@{}}
\toprule
Approaches / LP & Hi-Pa & Hi-Bn & Hi-Gu & Hi-Mr & Hi-Ta & Hi-Te & Hi-Ml \\ \midrule
NMT-BPE Baseline & 62.79 & 28.75 & 52.17 & 31.66 & 13.78 & 19.18 & 10.4 \\
Cognate-aware NMT-BPE & \textbf{65.55} & \textbf{29.43} & \textbf{52.39} & \textbf{32.41} & \textbf{13.85} & \textbf{19.58} & \textbf{11.18} \\
\bottomrule
\end{tabular}%
}
\caption{Results of the Cognate-aware Neural Machine Translation Task, in terms of BLEU scores, for the language pairs (LP) with available parallel data.}
\label{tab:mtresults}
\end{wraptable}
\renewcommand{\tabcolsep}{0.3cm}

The average improvement observed by using our best model (MUSE + WLS) over the strongest baseline approach~\cite{kanojia2019utilizing} is 9\% points with the highest being 18\% points (Hi-Ta). Over the weakest baseline approach (WLS), our best model obtains an average improvement of 50\%, peaking at 61\% points (Hi-Or).

We present the results of Cognate-aware NMT in Table \ref{tab:mtresults}. For the Hi-Pa language pair, an improvement of 2.76 BLEU is observed, where 15001 cognates were detected including the misclassified pairs. Amongst a consistent improvement for all the language pairs, even when 930 cognate pairs (Hi-Te) are added, an improvement of 0.4 BLEU can be seen. The maximum number of cognate pairs injected into the NMT pipeline is 15834 pairs for the Hi-Mr language pair. Surprisingly, we do not observe the most significant improvement for Hi-Mr despite the largest number of cognates injected. We believe that this is because Marathi is a morphologically rich language which exhibits agglutination.

\section{Conclusion and Future Work}
\label{sec:conc}

In this paper, we harness cross-lingual embeddings to improve the task of cognate detection for thirteen Indian language pairs. We propose the use of a linked knowledge graph to augment a publicly released cognate dataset with a context dictionary. We reproduce the proposed approach and add two additional language pairs to the same dataset and perform experiments using various approaches for a comparative evaluation. We reproduce the previously proposed approaches~\cite{rama2016siamese,kanojia2019utilizing} for this task to perform a further evaluation. We obtain monolingual Indian language corpora for all the fourteen languages (Section \ref{sec:des}), from various sources to build monolingual models and use a bilingual dictionary to supervise the task of cross-lingual models generation (MUSE and VecMap), for thirteen language pairs (Hi-Mr, Hi-Ta and so on). We also train a single cross-lingual model using the contextual embedding based approach (XLM-R). 

Our experiments use three different approaches to generate better feature representations for the cognate detection task, and all of them show improvements over previously proposed approaches. We observe consistent improvements in terms of precision, recall and F-scores. We also perform an ablation study and show that augmenting WLS baseline feature with MUSE based features provide us with the best results. Over the strongest baseline, this model shows improvements up to 18\% points, in terms of F-score.  Our best F-score is observed for the Hi-Gu and Hi-Ne language pairs (0.84) which can still be improved and warrants further investigation into the task. Additionally, we use the detected cognate pairs and use a simple approach to inject them into the neural machine translation pipeline. Our Cognate-aware NMT-BPE results also show a consistent improvement for all the Indian language pairs. Furthermore, we release this augmented dataset, along with our code and cross-lingual models for further research.

In future, we aim further to investigate the performance of contextual embeddings for this task. Recent trends show that contextual embeddings based models outperform conventional word embeddings for most tasks. We, however, do not observe this and attribute this primarily due to the dataset size used to train the contextual embeddings. We shall add more data to our monolingual corpora and perform more experiments using XLM-R. Future experiments with cognate-aware NMT using the Transformer architecture \cite{10.5555/3295222.3295349} should further help in showing the importance of our extracted cognate pairs. We also aim to investigate the task of cognate detection for other Indian language pairs, along with Indo-European language pairs, in the near future.

\section*{Acknowledgements}

We thank the lexicographers and annotators at the CFILT Lab, IIT Bombay for their efforts in creating the dataset for this study. We acknowledge the computational resources provided by NLP Lab at Monash University, and CFILT, IIT Bombay for performing the experiments. We also thank all the reviewers for their critique, which helped us shape up the article.

\bibliographystyle{coling}
\bibliography{coling2020}

\end{document}